\documentclass[runningheads]{llncs}
\usepackage[T1]{fontenc}
\usepackage{graphicx}
\usepackage{todonotes}
\usepackage{xcolor}
\usepackage{soul}
\usepackage{algorithm2e}
\usepackage{amsmath}
\usepackage{subcaption}
\usepackage{pdfpages}

\begin{document}
\title{Large Language Models for Explainable Decisions in Dynamic Digital Twins}
%
%
\author{Nan Zhang \inst{1}, Christian Vergara-Marcillo \inst{1,2}, Georgios Diamantopoulos\inst{1,2}, Jingran Shen\inst{1}, Nikos Tziritas \inst{3} \and Rami Bahsoon \inst{2} \and Georgios Theodoropoulos\thanks{Corresponding author}\inst{4,1}}
\authorrunning{N. Zhang et al.}
%
\institute{Southern University of Science and Technology, Shenzhen, China \and University of Birmingham, Birmingham, UK \and University of Thessaly, Lamia, Greece \and Research Institute of Trustworthy Autonomous Systems (RITAS), Shenzhen, China}
  \maketitle              
%


\begin{abstract}
Dynamic data-driven Digital Twins (DDTs) can enable informed decision-making and provide an optimisation platform for the underlying system. By leveraging principles of Dynamic Data-Driven Applications Systems (DDDAS), DDTs can formulate computational modalities for feedback loops, model updates and decision-making, including autonomous ones.  However, understanding autonomous decision-making often requires technical and domain-specific knowledge. 
This paper explores using large language models (LLMs) to provide an explainability platform for DDTs, generating natural language explanations of the system’s decision-making by leveraging domain-specific knowledge bases. A case study from smart agriculture is presented. 
\end{abstract}

\setcounter{footnote}{0} 

\section{Introduction}
\vspace{-.2cm}

\textit{Dynamic data-driven Digital Twins} (DDTs) utilise a DDDAS-based info-symbiotic feedback loop between a physical system and its virtual counterpart, whereby real-time data from the physical system are utilised to improve the system model, reconfigure the instrumentation pipeline, and optimise the underlying system. 
Due to the complexity and interdisciplinary nature of DDDAS, understanding the reasoning behind decisions requires vast technical and domain-specific knowledge. Furthermore, decisions are often represented by formats containing sensor readings and measurements that are hard for humans to parse. These challenges are exacerbated in the context of intelligent DDTs for intelligent systems, where DDTs can utilise their cognitive capabilities to make decisions autonomously \cite{9283357}. In this context, explainability becomes a critical property, allowing end-users to understand and, thus, trust the autonomous decisions of the DDT.


In \cite{zhang_explainable_2024}, we examined the concept and challenges of explainability in DDDAS/ DDT systems, identifying areas and stages in the analytics cycle that lend themselves to explainability. The aim of this paper is to extend that work by examining the role and utilisation of generative pre-trained large language models (LLMs) as a mechanism to support explainability. LLMs now possess the ability to mimic human-generated text with impressive accuracy, while through Retrieval-Augmented Generation (RAG) \cite{lewis_retrieval-augmented_2020}, LLMs can utilise knowledge bases (technical documents, scientific papers, computer code, etc.) when generating text. In the context of DDT explainability, RAG is the key feature, allowing LLMs to possess the required domain-specific knowledge to parse and sufficiently explain the decisions of the DDT. The contributions of the paper are the following: (a) An elaboration of different roles of LLM in supporting explainability in DDTs; (b) A novel reference architecture for LLM-enabled explainability for decisions in DDTs; (c) A proof-of-concept demonstration in a smart agriculture drone surveillance example for using LLM for explaining what-if simulation-based decisions.
 

The rest of the paper is structured as follows: Section \ref{sec:related-work} presents the related work in the explainable DDDAS/DDTs literature. Section \ref{sec:arch} introduces the proposed explainable DDT architecture. Section \ref{sec:example} presents an illustrative example of the proposed architecture in smart agriculture, including current limitations and possible future work. Finally, Section \ref{sec:conclusion} concludes this paper.

\vspace{-.3cm}
\section{Related Work} \label{sec:related-work}
\vspace{-.2cm}


Preliminary research has discussed explainability in DDTs. 
Intuitively, a system is interpretable if it uses a white box model based on physical or mechanical principles \cite{vilar-dias_interpretable_2024}. 
When black box models are needed, Explainable Artificial Intelligence (XAI) techniques are preferred \cite{zhang_explainable_2024}. 
Some XAI-empowered Digital Twins (DTs) focus on explaining the predictions, such as time-series forecast \cite{neupane_twinexplainer_2023}, reliability of prediction and model retraining \cite{an_explainable_2023}.
Some others relate to explaining the autonomous behaviours of DDT, such as dynamically selecting a model from a model library to estimate structural health \cite{kapteyn_toward_2020}.
Architectural efforts to enhance DT explainability include a three-tiered pipeline from \cite{neupane_twinexplainer_2023} for explaining DT predictions and a self-explainability framework in \cite{michael_explaining_2024} where the explanation model is derived from formal system specifications.
Explanation during design also enhances understanding among modellers and stakeholders \cite{wang_explainable_2021}.
Despite the work in enriching DTs with explainability, the focus on explainable \textit{decisions} autonomously made by the DDT is still at the preliminary stage \cite{zhang_explainable_2024,kapteyn_toward_2020,michael_explaining_2024}.


LLMs are increasingly augmenting scientific simulations, offering potential in elucidating model structures, summarising simulation outputs, enhancing platform accessibility, and interpreting errors \cite{giabbanelli_gpt-based_2024}.
An individual LLM agent can understand and act on human natural language-based instructions to perform remote simulations \cite{cao2024llmassisted}. Extending this, multiple LLM agents are being explored for advanced tasks, including simulating human behaviours \cite{yang_llm-based_2024} and participating in complex processes like perception, planning, and decision-making \cite{sun_empowering_2024}. 
Despite general explorations into LLM explainability \cite{zhao_explainability_2024}, the specific roles of LLMs in explaining DDDAS/DDTs require further clarification.
Rather than using LLM for decision-making, this paper adopts LLMs as an intermediate layer between the DDT and humans for organic integration between trustworthy simulation analysis techniques and LLM explanations. 


\section{A Reference Architecture} \label{sec:arch}

\begin{figure}[t]
\vspace{-.4cm}
\centering
\includegraphics[width=1\textwidth]{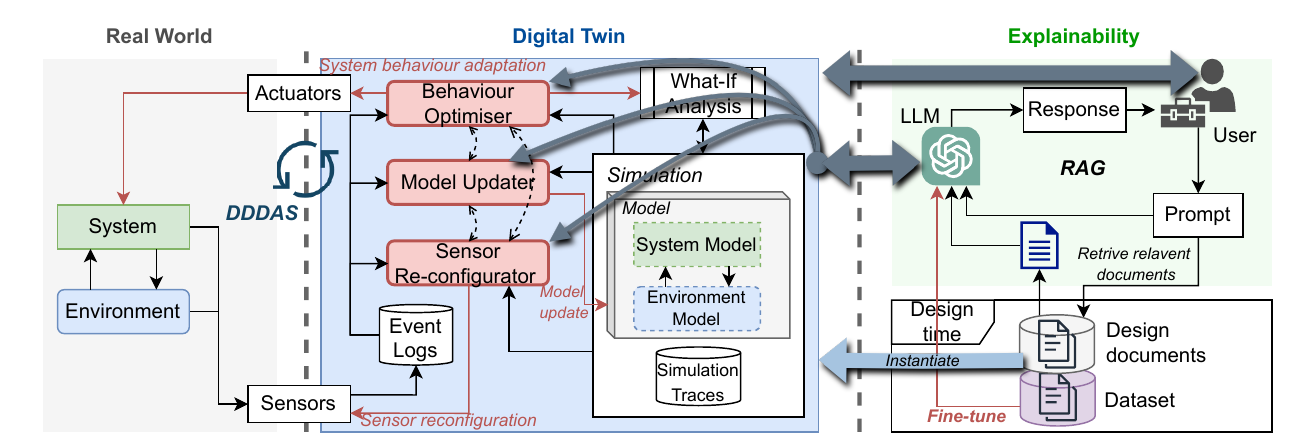}
\caption{LLM enabled explainability in a DDT system.} \label{fig:arch}
\vspace{-.5cm}
\end{figure}

In \cite{zhang_explainable_2024} we presented a reference architecture that illustrated the explainability requirements in DDTs. Based on that architecture, Fig. \ref{fig:arch} illustrates the potential role of LLM in supporting explainable decisions in DDT.   The architecture considers LLM's role in supporting explainability in different dimensions: who makes decisions, how the decisions are explained, and how humans are involved.
In \cite{zhang_explainable_2024} three types of explainable decisions were identified, leveraging the classical DDDAS architecture \cite{blasch_dddas_2018}: the explanation for \textit{measurement adaptation}, \textit{model adaptation}, and \textit{system behaviour adaptation}.
It can be argued that LLM can further support these three types by enhancing the understandability and the ease of human-system interaction and value alignment.

\vspace{-.5cm}
\subsection{Roles of LLM for Explainability}
\vspace{-.2cm}

LLM acts as a bridge between offline design and real-time data, employing specialised domain-related knowledge. It offers customisable explanations for diverse stakeholders through fine-tuning, ensuring reliable performance. For DDT in particular, the role of LLM varies depending on who the decision maker is. This paper considers the following four scenarios:

\textbf{DDDAS/DDT makes decisions:}
LLM should serve as an external interpreter outside of the DDT to interact with the users by elucidating the decision-making process, inputs, and decisions, without intruding on the DDT's operations. It achieves this by analysing the DDT system's design documentation, inputs, and operational logs. 

\textbf{Human makes decisions:}
Human stakeholders require comprehensive information to make informed decisions, yet their understanding of DDT can vary significantly. They need to understand the state space, control space, objective space, and the mapping between them, enabled by techniques such as what-if analysis. 
LLMs, in this case, must not only interpret the DDT's behaviour but also facilitate a two-way dialogue between humans and the DDT. The LLM should:
1) clarify the current system state to humans, 2) translate human natural language-based instructions into machine-readable commands for the DDT to perform simulation analyses, 3) communicate and explain the results of these simulations, 4) convey human decisions back to the DDT, potentially suggesting alternative decisions.
    
\textbf{LLM makes decisions:}
A planner LLM operates as a black box, requiring transparent explanations of its decision-making. It monitors the system, creates prompts for itself, and devises plans. To enhance transparency and decision accuracy, Chain-of-Thought prompting can be employed to articulate the decision rationale step-by-step \cite{NEURIPS2022_CoT}. Yet, the LLM's generative and probabilistic nature can make trusting its explanations difficult, referred to as hallucination \cite{huang_survey_2023}.

\textbf{Collaborative decision-making:}
Integrating the three aforementioned scenarios, a collaborative framework can be established, involving DDDAS, LLMs, and humans. Each entity can take on different decision-making responsibilities and provide feedback to enhance collective decision quality. While DDDAS may face constraints leading to sub-optimal decisions, LLMs can offer reassessments and recommend human interventions. Similarly, LLMs can critique and refine human decisions, offering insights into potential enhancements.

The rest of the paper specifically investigates one specific scenario above where the DDT is the decision-maker, with the LLM explaining the DDT's decisions on \textit{system behaviour adaptation} to optimise task-related goals.

\vspace{-.4cm}
\section{An Illustrative Example: Drone Fleet in Smart Farming} \label{sec:example}
\vspace{-.2cm}

\begin{figure}
\centering
\vspace{-.3cm}
\includegraphics[width=1\textwidth]{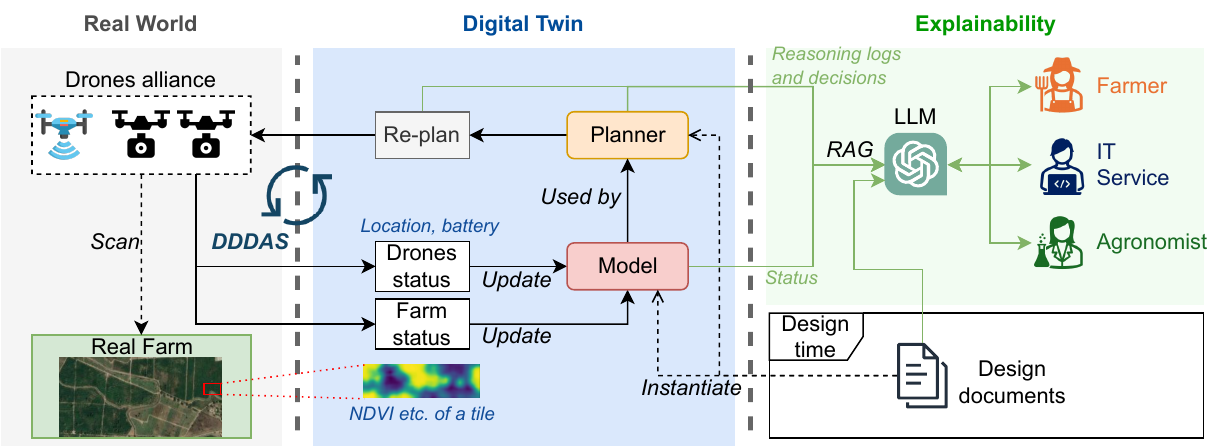}
\caption{The DDDAS loop of the DDT in the smart farming application scenario.} \label{llm-app}
\vspace{-.5cm}
\end{figure}

Smart farming applications can be assisted by aerial monitoring using drones and remote sensing to detect problems assessing crop health through vegetation indices such as NDVI \cite{zhang2014crop}. Nevertheless, these drone-based systems are often time-intensive, leading to challenges in sensing and model updates for achieving efficient sensing and prompt updates to models, further constrained by factors such as energy consumption and weather. Given that the system's status changes in real-time, continuous model updates are necessary. A DDT-based approach can dynamically adjust drones' sensing behaviour 
enabling time and cost-efficient updates. Previous papers have considered DDT-based approaches for smart agriculture \cite{9361651,8642114,9004854,10.1007/978-981-15-1078-6_1}. This paper adopts a DDT-enabled smart farming application to showcase the benefits of integrating LLMs and DDDAS in a multi-resolution application made of a drone fleet or \textit{alliance}. In this case, two types are proposed: Survey and Inspection drones. The former facilitates a quick scan at high altitudes but with low resolution, mainly used to identify Regions Of Interest (ROIs). The latter performs a focused exploration of these ROIs. DDT, in this scenario, enables dynamic drone planning designed to enhance efficiency in the overall system's operations.
The full workflow is shown in Fig. \ref{llm-app}.

\vspace{-.2cm}
\subsection{Modelling}

\RestyleAlgo{ruled}

\begin{figure}[htbp]
    \vspace{-.6cm}
    \centering
    \resizebox{0.85\linewidth}{!}{
        \begin{algorithm}[H]
            \caption{The planning workflow}\label{algorithm}
            A new image is received from the survey drone\;
            Calculate its confidence\;
            \If{confidence < $T_{\alpha}$}
            {
                $R \gets [\ ]$\;
                \tcc{Assign an inspection drone to examine it}
                \For{d in all inspection drones}{
                    $\Delta t, battery \gets Simulate(d)$\;
                    \If{battery > $T_b$}
                    {
                        $R.append(\Delta t)$\;
                    }
                }
                $d_o \gets$ rank all options in $R$ and get the drone with the minimum time $\Delta t$\; 
                Send instructions to $d_o$\;
            }
            \label{alg:dddas_planner}
        \end{algorithm}
    }
    \vspace{-.6cm}
\end{figure}

\noindent\textbf{Farm model:} The farm can be envisioned as a grid of \textit{tiles} of the same size.
The model of the farm provides the healthiness of each tile, built by images with NDVI taken by the drone alliance in real-time. Farmers or automated equipment will use this information to apply the \textit{exact} amount of nutrients or water. 

Due to the information loss in the lower-resolution images taken by the survey drone, a confidence indicator $\alpha_i$ of a tile $i$ is introduced to decide whether there is a need to let inspection drones examine carefully. An intuitive idea is to measure the confidence that the crops in the tile are healthy or unhealthy. The confidence is low if it is difficult to tell whether the tile is definitely $Good$ or $Bad$.
\begin{equation}
    \alpha_i = | \mu_{Good}(\bar{V}_i) - \mu_{Bad}(\bar{V}_i) | \quad \textrm{and} \quad
    \bar{V}_i = \frac{1}{N} \sum_j^{N} V_{NDVI,j}
\end{equation}
where $\bar{V}_i \in [-1,1]$ is the average NDVI of all pixels of the tile $i$; 
$\mu$ is a fuzzy membership function shown below to define the degree that the health condition of the crop is $Good$ or $Bad$ ($a_1, b_1, a_2, b_2 \in [-1,1]$ are parameters to define the boundary of the membership).
\begin{equation}
    \resizebox{.7\linewidth}{!}{$
    \begin{aligned}
        \mu_{Good}(x) &=
        \begin{cases}
          0 & \text{$x < a_1$}\\
          \frac{a_1-x}{a_1-b_1} & \text{$a_1 \leq x \leq b_1$}\\
          1 & \text{$x > b_1$}
        \end{cases}
        ,\quad
        \mu_{Bad}(x) &=
        \begin{cases}
          1 & \text{$x < a_2$}\\
          \frac{a_2-x}{a_2-b_2} & \text{$a_2 \leq x \leq b_2$}\\
          0 & \text{$x > b_2$}
        \end{cases}
    \end{aligned}
    $}
\end{equation}

\vspace{.2cm}
\noindent\textbf{Drone model:} The drone model considers the flying time and remaining battery life. 
If a drone is busy inspecting a tile and receives an instruction to inspect another tile located at $s$ unit distance, it will first finish the current task, which will take $t_{rem}$ time units. Then, the drone will fly to the new tile at speed $v$ ($t_{disp} = s/v$) and will proceed to inspect it ($t_{insp}$). Therefore, the total time for inspecting a new tile is: 
$\Delta t = t_{rem} + t_{disp} + t_{insp}$.

\vspace{.2cm}

\noindent\textbf{Planning model:} The re-planning of drones is based on the status of all drones and tiles.
A tile requires further inspection if its confidence $\alpha_i$ is less than a threshold $T_\alpha$.
What-if analysis is conducted to simulate the options for sending a particular inspection drone. 
Only drones with enough predicted remaining battery life (larger than $T_b$) to finish the task are potential candidates.
All candidate options are then sorted by the total time, and the drone with the minimum time $\Delta t$ is selected.
The detailed workflow is shown in Algorithm \ref{algorithm}.

\vspace{-.3cm}
\subsection{LLM-Enabled Explainability}

\begin{figure}
\vspace{-.4cm}
    \centering
    \includegraphics[width=\textwidth]{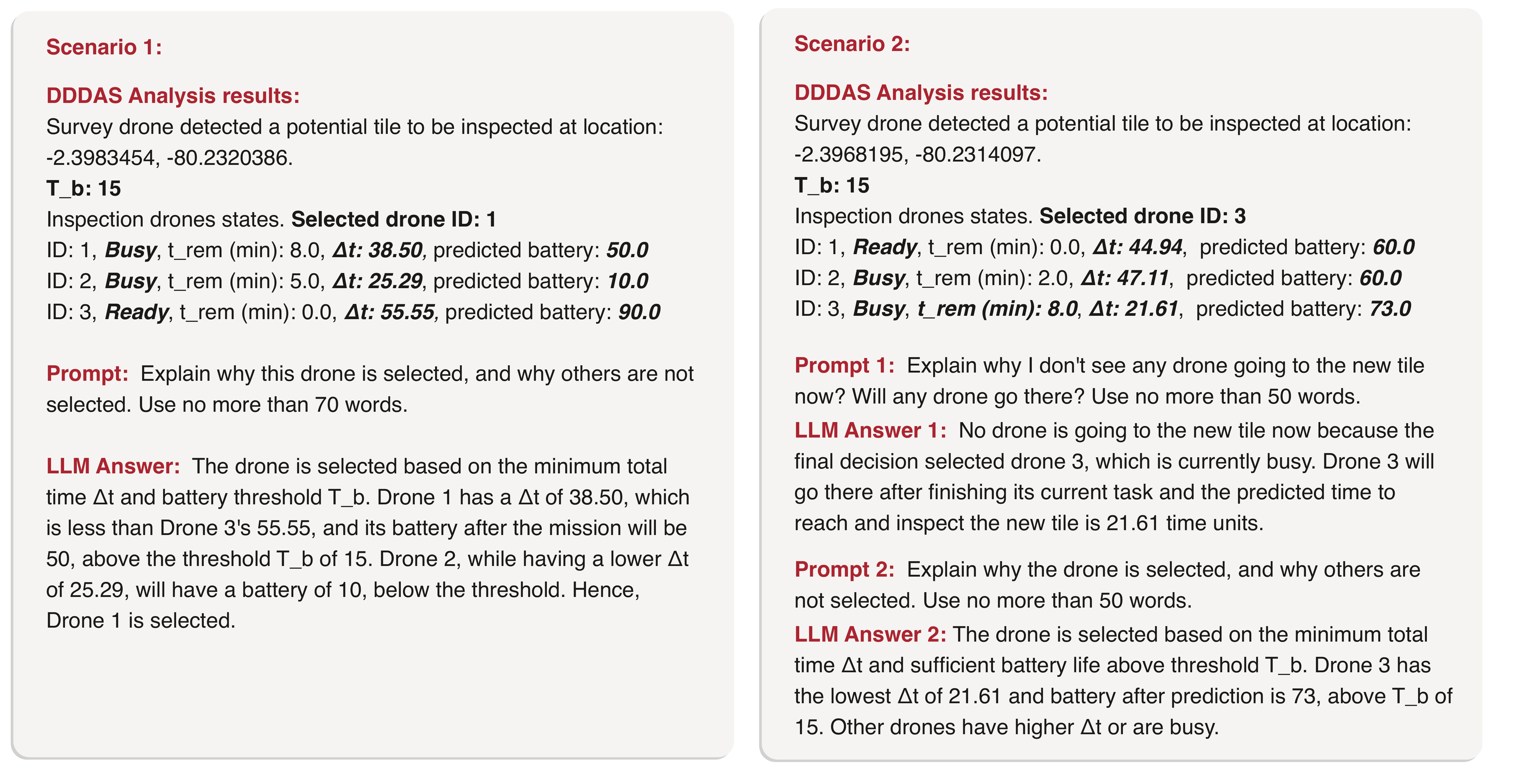}
    \vspace{-.5cm}
    \caption{DDDAS/DDT decision-making and LLM's explanation with RAG.}
    \label{fig:dddas-llm}
\vspace{-.5cm}
\end{figure}

The complexity of the drone system's dynamic planning, as outlined in Algorithm~\ref{alg:dddas_planner}, can lead to seemingly counter-intuitive actions, necessitating clear explanations for farm operators. LLMs, with their vast public data training, are ideal for this but lack access to specific or local knowledge. Therefore, Retrieval Augmented Generation (RAG) is used to integrate a custom knowledge base, ensuring LLMs provide coherent and contextually relevant information \cite{lewis_retrieval-augmented_2020}. 
This study integrates RAG with existing LLM APIs and domain expertise through tools like LangChain to optimise response quality.

To demonstrate the LLM's explainability capabilities, we conducted experiments\footnote{Code available at: https://github.com/explainable-digital-twins/RAG-DDDAS} using  LangChain, OpenAI \verb|gpt-4-turbo-2024–04–09| model and embedding model \verb|text-embedding-ada-002|,
providing the LLM with contextual information and decisions made by the DDDAS loop of the DDT. Relevant runtime information includes drones status, and 
structured DDDAS/DDT simulation results (in JSON) regarding time and battery to complete an emergent task. 
The RAG knowledge base includes the content of this paper before this section (as a \verb|.tex| text file) and an additional description of the JSON results.
Two scenarios are considered in Fig. \ref{fig:dddas-llm}.
In \textbf{Scenario 1} the DDT instructs a drone (Drone 1) whose total predicted time $\Delta t$ is not the minimum among all drones, because the minimum one (Drone 2) does not meet the battery requirement of $T_b$. 
Fig. \ref{fig:dddas-llm} shows that the LLM successfully link the value $T_b$ (denoted as \verb|T_b| and with no further information provided in the simulation results) with the definition in the planning model by explicitly emphasising the relationship between the threshold and the remaining battery life.
In \textbf{Scenario 2}, 
the DDT instructs a ``busy'' drone to finish its current task first and then inspect a new tile because dispatching a new ``ready'' drone (awaiting instructions) takes more time. Users can be confused as to why no drone was immediately dispatched for this task. However, the LLM explains the reason for this context-related question as in \verb|LLM Answer 1|. In \verb|LLM Answer 2|, the LLM also clearly explains the role of $T_b$ and the decision-making process adhering to the planning model specified in the previous subsection.

This demonstration showcases a promising vision for LLM to enhance the explainability of the system. 
The LLM is able to extract and synthesise relevant information from different types of data according to unstructured input questions, and generate explanations that require reasoning.
However, the hallucination problem can be challenging as LLMs can generate irrelevant or wrong information \cite{huang_survey_2023}.
For instance, ``\textit{Other drones have higher $\Delta t$ or are busy}'' appears in \verb|LLM Answer 2| of Fig. \ref{fig:dddas-llm}: Scenario 2, but whether the drone is busy is not considered in the planning algorithm, which requires further tuning of the prompt and retrieval process.
LLM is also powerful for providing explanations tailored specifically for users with different knowledge backgrounds, such as farmers, IT support teams, and agronomists, as shown in Fig. \ref{llm-app}. Future work will focus on introducing user models for specialised explanations.

\section{Conclusion} \label{sec:conclusion}
This paper investigates the role of LLMs in enhancing explainability for \textit{Dynamic data-driven Digital Twins} (DDTs). Through a smart agriculture case study, we have demonstrated with RAG how LLMs can provide natural language explanations for autonomous decisions made by the DDT. 
Despite challenges, including the risk of hallucination, the integration of LLMs with DDT holds significant potential, paving the way for more trustworthy autonomous systems across domains.
More sophisticated scenarios should be evaluated.
Future research should also consider i) fine-tuning the LLM with domain datasets, integrating the LLM in a complete workflow with a running simulator and field deployment, and ii) including a systematic validation of the generated explanation, possibly with the involvement of human experts.
\begin{credits}
\subsubsection{\ackname} This study was supported by: Shenzhen Science and Technology Program, China (No. GJHZ20210705141807022); SUSTech-University of Birmingham Collaborative PhD Programme; SUSTech Research Institute for Trustworthy Autonomous Systems, Shenzhen, China.

\end{credits}

%
%
%
\bibliographystyle{splncs04}
\bibliography{mybibliography}

\end{document}